\documentclass[conference]{IEEEtran}
\IEEEoverridecommandlockouts
\usepackage{cite}
\usepackage{amsmath,amssymb,amsfonts}
\usepackage{algorithmic}
\usepackage{graphicx}
\usepackage{textcomp}
\usepackage{xcolor}
\usepackage{bbm}

\usepackage{amsfonts}
\usepackage{booktabs}
\usepackage{siunitx}
\usepackage[T1]{fontenc}
\usepackage[utf8]{inputenc}
\usepackage{multirow}
\usepackage{enumitem}

\usepackage[ruled,vlined, linesnumbered]{algorithm2e}

\usepackage[english]{babel}
\usepackage[bookmarks=false]{hyperref}
\addto\extrasenglish{%
}

\addto\extrasenglish{ 	 	 	 }

\addto\extrasenglish{%
}
\usepackage{adjustbox}
\newcommand{\ours}{DAM3}

\usepackage{float}

\usepackage{todonotes}
\usepackage{afterpage}

\def\BibTeX{{\rm B\kern-.05em{\sc i\kern-.025em b}\kern-.08em
    T\kern-.1667em\lower.7ex\hbox{E}\kern-.125emX}}

\makeatletter
    
\begin{document}

\title{Drift-Aware Multi-Memory Model for \\Imbalanced Data Streams}

\author{
\IEEEauthorblockN{Amir Abolfazli}
\IEEEauthorblockA{\textit{L3S Research Center}\\
\textit{Leibniz University of Hanover}\\
Germany\\
abolfazli@L3S.de}
\and
\IEEEauthorblockN{Eirini Ntoutsi}
\IEEEauthorblockA{\textit{L3S Research Center}\\
\textit{Leibniz University of Hanover}\\ 
Germany\\
ntoutsi@L3S.de}
}

\maketitle

\begin{abstract}
Online class imbalance learning deals with data streams that are affected by both concept drift and class imbalance. Online learning tries to find a trade-off between exploiting previously learned information and incorporating new information into the model. This requires both the incremental update of the model and the ability to unlearn outdated information. The improper use of unlearning, however, can lead to the \emph{retroactive interference} problem, a phenomenon that occurs when newly learned information interferes with the old information and impedes the recall of previously learned information. The problem becomes more severe when the classes are not equally represented, resulting in the removal of minority information from the model. 
In this work, we propose the Drift-Aware Multi-Memory Model (DAM3), which addresses the class imbalance problem in online learning for \emph{memory-based models}. DAM3 mitigates class imbalance by incorporating an imbalance-sensitive drift detector, preserving a balanced representation of classes in the model, and resolving retroactive interference using a working memory that prevents the forgetting of old information. We show through experiments on real-world and synthetic datasets that the proposed method mitigates class imbalance and outperforms the state-of-the-art methods.
\end{abstract}


\section{Introduction}
\label{sec:intro}
The challenge of learning from imbalanced data streams with concept drift has attracted a lot of attention from both academia and  industry in recent years. The term \emph{concept
drift} refers to changes in the underlying data distribution over time. \emph{Class imbalance} occurs when the classes are not represented equally. Online class imbalance learning deals with data streams that are affected by both concept drift and class imbalance and exists in many real-world applications such as anomaly detection, risk management, and social media.

Online learning algorithms, dealing with imbalanced streams, not only try to better represent the minority class for the learning model (e.g., by oversampling the minority class) but also try to find a trade-off between retaining the previously learned information and adapting to new information from the stream, known as the \emph{stability-plasticity
dilemma} \cite{carpenter1987art}. The basic idea is that a learning model requires \emph{plasticity} for the integration of new information, but also \emph{stability} in order to prevent the forgetting of old information \cite{mermillod2013stability}. A too adaptive model forgets previously learned information and a too stable model cannot learn new information, hence, finding a trade-off between plasticity and stability is required.

In recent years, many methods have been proposed to deal with online class imbalance learning (e.g., \cite{wang2016online, lu2017dynamic, zhang2019resample}). Some methods employed \emph{unlearning} to deal with concept drift (e.g., \cite{krawczyk2015weighted, losing2016knn, yu2019adaptive}) by removing the information that is inconsistent, i.e., contradicting class labels, with the incoming data from the stream. SAM-kNN~\cite{losing2016knn} is one of such models based on kNNs that makes use of unlearning in the neighborhood of the instances. SAM-kNN is a dual-memory model~\cite{losing2016knn} that partitions the knowledge between short-term memory (STM) and long-term memory (LTM), containing the information of the current and former concepts, respectively. Preserving the consistency, in SAM-kNN, is based on a cleaning operation that unlearns the information of former concepts in the LTM that contradicts the information of the most recent concept, stored in the STM. Although \emph{unlearning} is a desired property for the model adaptation, if not applied carefully, it could lead to the \emph{retroactive interference} problem \cite{sosic2018learning} that occurs when new information interferes with previously learned information, causing the (unintentional) forgetting of old information.

\autoref{fig:retroactive_interference} illustrates the problem of retroactive interference in a dual-memory model, where old information in the LTM is removed because the model adapts to new data in the STM. The problem becomes more severe when the classes are not equally represented in the stream as it could lead to the removal of minority instances, which are of higher interest than majority instances in many real-world applications.

In this work, we propose a multi-memory model which deals with the class imbalance by 1) incorporating an imbalance-sensitive drift detector, 2) preserving a balanced representation of classes in the model, and 3) resolving the retroactive interference by means of a \emph{working memory} (WM)~\cite{diamond2013executive}, manipulating information in the LTM for every incoming instance to the STM. We also contribute new synthetic benchmarks with different drift types and class imbalance ratios.

\afterpage{%
\begin{figure}[!h]
\center
\includegraphics[width=0.44\textwidth]{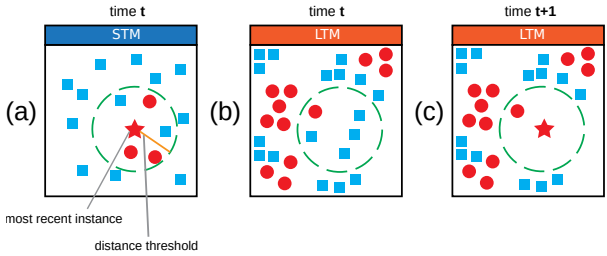}
\caption{Illustration of retroactive interference problem in a dual-memory model. \textbf{(a)} the red star instance is the most recent instance in the STM. Its neighborhood (indicated by the green dashed circle) is defined by the maximum distance of consistent instances (i.e., instances with the same class: red) in its neighborhood ($k=5$). \textbf{(b}) state of the LTM at $t$, the affected area (indicated by the green dashed circle) centered at the new point's location. 
\textbf{(c)} state of the LTM at $t+1$, where (the majority of) the previously stored instances have been removed due to inconsistency with the new information of the STM.}
\label{fig:retroactive_interference}
\end{figure}
}
\section{Preliminaries and basic concepts} \label{sec:imbalanced_stream_classification}
A data stream $D$ is a potentially infinite sequence of instances arriving at distinct time points $1, \cdots, t, \cdots$, where $t$ is the current timepoint. Each instance $\mathbf{x} \in D$ is described in a $d$-dimensional feature space, i.e., $x \in \mathbb{R}^{d}$. Without loss of generality, we assume a binary classification problem, i.e., $Y = \{+, -\}$. We follow the  first-test-then-train or prequential evaluation setup~\cite{gama2014survey}. Assuming a probability distribution $P(\mathbf{x},y)$ generating the instances of $D$, the characteristics of $P$ might change with time, i.e.,
for two time-points $i$, $j$, it might hold that $P_i(\mathbf{x},y) \neq  P_j(\mathbf{x},y)$, a phenomenon called \emph{concept drift}~\cite{gama2014survey}. The drift type could also be characterized based on the rate at which drift occurs. \emph{Sudden drift} results in a severe change in the distribution of data. \emph{Incremental drift} occurs when the concept incrementally changes. \emph{Gradual drift} occurs when the instances belonging to two different concepts are interleaved for a certain period of time.
\emph{Recurring drift} describes a case in which the concept which has already been observed, reoccurs. Apart from the occurrence of concept drifts, we also assume that the stream is \emph{imbalanced} with the majority class (assumed to be $-$) occurring more often than the minority class (assumed to be $+$). To express the degree of imbalance, we use the \emph{imbalance ratio (IR)}~\cite{orriols2009evolutionary} defined as the number of minority instances over majority instances. IR is commonly denoted by 1:r (r is a value corresponding to the majority class) which specifies the ratio between the minority and majority class~\cite{fernandez2018learning}. In a streaming data environment, imbalance can be either \emph{static} assuming a fixed class ratio or \emph{dynamic} assuming varying class ratio over the stream. Learning under imbalance is harder in a stream environment as there is no prior knowledge about the IR and often the role of minority and majority changes over the stream~\cite{ntoutsi2012density}.
\section{Related Work}
\label{sec:related_work}
Online class imbalance learning methods deal with data streams, affected by both concept drift and class imbalance.

Bifet et al. in \cite{bifet2009new} proposed OBA which is an online ensemble method that improves the Online Bagging algorithm \cite{oza2005online} by adding the ADWIN change detector. When a change is detected, the worst-performing base learner of the ensemble is replaced with a new one. Bifet et al. in \cite{bifet2010leveraging} proposed Leverage Bagging (LB), an online ensemble method that leverages the performance of bagging by increasing the weights of the resampling using a larger value $\lambda$ to compute the value of the Poisson distribution to increase diversity of the ensemble.
 
LB uses the ADWIN change detector to deal with concept drifts. When a concept drift is detected, the worst base learner is reset. Both OBA and LB indirectly deal with class imbalance as Diez-Pastor et al. in~\cite{diez2015diversity} showed that diversity-increasing techniques such as bagging improve the performance of ensemble methods for imbalanced problems.

Melidis et al. in \cite{melidis2018learning} proposed a method assuming the likelihood of different features with respect to the class follows different trends and proposed an ensemble method that predicts the best trend detector.

Wang and Pineau in \cite{wang2016online} proposed online AdaC2, an online boosting algorithm that considers the different misclassification costs when calculating the weights of base learners and updates the weights of instances accordingly. More precisely, AdaC2 increases weight more on the misclassified positive instances than the misclassified negative instances. The same authors in \cite{wang2016online} proposed the online RUSBoost, an online boosting algorithm that removes instances from the majority class by randomly undersampling the majority-class instances in each boosting round. The original version of both AdaC2 and RUSBoost do not deal with concept drift. However, the improved version of these methods deal with concept drifts using an ADWIN change detector.

The most relevant work to our work is the Self Adjusting Memory model for the $k$ Nearest Neighbor algorithm (SAM-kNN)~\cite{losing2016knn}. SAM-kNN builds an ensemble of classifiers induced on different memories: the short-term memory (STM) for the current concept and the long-term memory (LTM) for the former concepts, and the combined memory (CM) which is the union of STM and LTM. The authors propose a cleaning operation during the transfer that deletes instances of the LTM that are inconsistent with transferred instances of the STM.

The original SAM-kNN model does not consider class imbalance. In case of imbalance, the memories and in particular the LTM memory is increasingly dominated by the majority-class instances (see \autoref{fig:weatherIR} (b)). As a result, the performance of the model on minority instances is dropping. As we show in our experiments, this is not only because of the reduced representation of the minority class in the input stream but also because of the cleaning operation which deletes more instances of the minority class (see \autoref{fig:weather_memoryInconsistency} (e)).

Unnikrishnan et al. in \cite{unnikrishnan2020entity} proposed specialized kNN models (for each entity) and a global kNN (for the whole stream) to ensure adequate representation of the entities in the learning models, independent of their volume. Moreover, their method leverages the global model to deal with the cold-start problem. 

Recently, the problem of fairness-aware learning in the online setting and under the class imbalance has been introduced~\cite{iosifidis2020fabboo}; the proposed solution changes the training distribution to take into account the evolving imbalance and discriminatory behavior of the model, both of which are evaluated over the historical stream.

\section{Drift-Aware Multi-Memory Model}
\label{sec:proposed_method}
In the research field of human memory, multi-memory models~\cite{atkinson1971control} have been proposed to overcome the limitations of dual-memory models and better represent the human memory. Such models consist of the sensory register (SR), short-term memory (STM), and long-term memory (LTM). The basic idea is that, first, the sensory information enters the SR, keeping the information for a very short time. The sensory information is transferred into the STM for temporary storage, and is encoded visually, acoustically or, semantically. The information is then transferred to the LTM after getting enough attention by processes such as active rehearsal~\cite{atkinson1971control}. The information that enters the STM is joined by context-relevant information in the LTM, which requires the retrieval of information from the LTM. Sometimes, the information of the LTM cannot be retrieved due to the \emph{retroactive interference (RI)}~\cite{sosic2018learning}. 
RI is the interference occurring when newly learned information impedes the recall of previously learned information. A theoretical concept proposed in the field of cognitive psychology is the \emph{working memory} \cite{baddeley1974working}, which introduces a memory that temporarily stores information relevant to the current task.

In a dual-memory model (see \autoref{fig:retroactive_interference}), the problem of retroactive interference occurs when inconsistent information of the LTM is replaced with new information of the STM. Such a replacement, intended for dealing with concept drifts in SAM-kNN~\cite{losing2016knn}, can result in loss of information that happens 1) when information is transferred from the STM to the LTM, and 2) when the LTM is cleaned with respect to the STM for every incoming instance from the stream.
As we will see in the experiments (\autoref{fig:weatherIR} (b)), such a replacement greatly affects the minority class, and therefore, the RI problem becomes more severe for the minority class.

To deal with this problem, we incorporate a \emph{working memory (WM)}, preventing the model from removing inconsistent instances, into our model. Due to the fact that stream contains more instances from the majority class and also the fact that the LTM is cleaned with respect to every incoming instance from the stream, more minority instances become inconsistent and thus removed from the LTM. Therefore, the minority class benefits more from the use of the WM. 

\begin{figure}[!t]
\center
\includegraphics[width=0.44\textwidth]{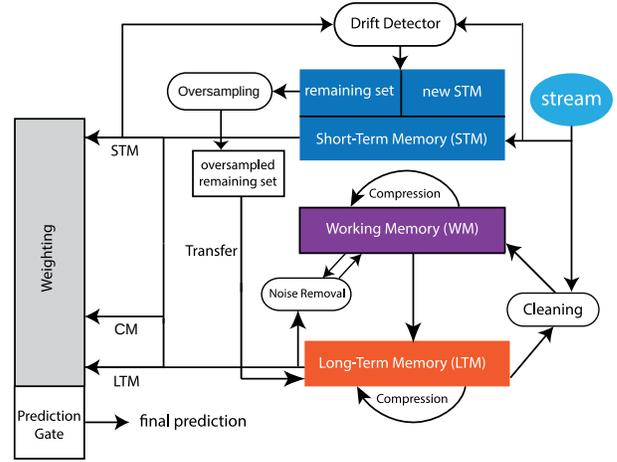}
\caption{Architecture of DAM3.}
\label{fig:architecture}
\end{figure}

The proposed model, DAM3, is a multi-memory model for data streams with class imbalance and concept drifts that: i) introduces a working memory to deal with the retroactive interference (\autoref{sec:our-memorymodels}), ii) incorporates an imbalance-sensitive drift detector (\autoref{sec:our-driftdetection}) to take into account the inherent imbalance, iii) preserves a balanced representation of classes using oversampling, (\autoref{sec:oversampling}), and iv) removes noisy instances in the working memory that are generated after exchanging information (\autoref{sec:noise-removal}). The architecture of DAM3 is shown in \autoref{fig:architecture}. 

\subsection{Model memories}
\label{sec:our-memorymodels}
DAM3 consists of four memories: $STM$, $WM$, $LTM$, and $CM$, each of which is represented by a set of labeled instances.

\emph{Short-term memory (STM)} is dedicated to the current concept and is a dynamic sliding window containing the most recent $m$ instances from the stream ($t$ is the current timepoint):
\[
STM=\left\{\left(\mathbf{x}_{i}, y_{i}\right) \in \mathbb{R}^{d} \times\{+1, -1\} | i=t-m+1, \ldots, t\right\}.
\]

\emph{Long-term memory (LTM)} maintains information ($p$ points) of former concepts that is \emph{consistent} with the current concept:
\[
LTM=\left\{\left(\mathbf{x}_{i}, y_{i}\right) \in \mathbb{R}^{d} \times\{+1, -1\} | i=1, \ldots, p\right\}.
\]

\emph{Combined memory CM (STM $\cup$ LTM)} represents the combination of short-term and long-term memories. It comprises just the union of STM and LTM and has the size $m+p$.

\emph{Working memory (WM)} lends itself to resolving the retroactive interference problem. It preserves inconsistent information of the LTM and also transfers back (to the LTM) information that becomes consistent with the most recently stored information in the STM. In this way, the WM makes its consistent information available to the LTM for current predictions, made by the classifiers LTM and CM, and also retains valuable information for later predictions. The WM is a set of $q$ points:
\[
WM=\left\{\left(\mathbf{x}_{i}, y_{i}\right) \in \mathbb{R}^{d} \times\{+1, -1\} | i=1, \ldots, q\right\}.
\]

\subsection{DAM3 training, weighting, and prediction}
\label{sec:model-training-prediction}
Each memory induces a classifier; therefore, DAM3 could be considered as an ensemble method. 
\subsubsection{DAM3 training}
In the SAM-kNN~\cite{losing2016knn} model, weighted kNN classifiers were employed for all memories. In this work, we use the weighted kNN for the \emph{LTM} and \emph{CM} as it allows the seamless implementation of cleaning operation.

kNN assigns a label to an instance $\mathbf{x}$ based on the memory instances:
\begin{equation}
\mathrm{kNN}_{M}(\mathrm{x})=\arg \max  _{\hat{y}} \sum_{x_{i} \in N_{k}(\mathbf{x}, M) | y_{i}=\hat{y}} \frac{1}{d\left(\mathbf{x}_{i}, \mathbf{x}\right)}
\end{equation}

where $d\left(\mathbf{x}_{1}, \mathbf{x}_{2}\right)$ is the Euclidean distance between two points, $N_{k}(\mathbf{x}, M)$ is the set of $k$ nearest neighbors of $\mathbf{x}$ in $M$, and $M \in \{LTM, CM\}$.

We use the \emph{full Bayes} classifier~\cite{berthold2010guide} for the \emph{STM} instead of the weighted kNN as the instance-based learning classifiers are quite sensitive to noisy data~\cite{qin2013cost}. In our case, the use of the kNN as the STM classifier might result in incorrect predictions passed to the drift detector (c.f., \autoref{sec:our-driftdetection}). 

The full Bayes classifier assumes that the distribution of data can be modeled with a multivariate Gaussian distribution~\cite{berthold2010guide}. A new instance $\mathbf{x}$ is classified as follows:
\begin{equation}
\operatorname{FB_{STM}}(\mathbf{x})=\arg \max _{y} p(y) f(\mathbf{x} | y)
\end{equation}
where $p(y)$ is the class prior and $f(\mathbf{x}|y)$ is the multivariate Gaussian density function~\cite{berthold2010guide}.

\subsubsection{DAM3 weighting and prediction}
Each of the base classifiers STM, LTM, and CM is weighted based on its balanced accuracy on the most recent $ms$ instances of the stream, where $ms$ is equal to the minimum size of the STM. The best performing model is chosen to predict the class label of the current instance.

\subsection{Imbalance-sensitive drift detection}
\label{sec:our-driftdetection}
Information is transferred from the STM to the LTM when concept changes. A change in concept is signaled by significant degradation of the performance of the STM classifier, corresponding to the model learned on the current concept. Due to inherent class imbalance of the stream, performance of the model on both classes should be taken into account. Hence, we propose a drift detector that relies on balanced accuracy. The detector takes as inputs the incoming instances and their corresponding predictions made by the STM classifier.

Our proposed drift detector divides all the balanced accuracy values into two windows (reference window and test window) and performs the non-parametric Kolmogorov–Smirnov test. If the balanced accuracy values of the reference window are significantly different from those of the test window, the drift is detected and the STM size is reduced from $m$ to $ws$, where the $m$ is the current size of the STM and $ws$ is the window size of the drift detector (each of the reference and test windows has the size $ws$), respectively. In this way, the test window becomes the new STM ($STM_{t+1}$, with the size $ws$) and the remaining set (denoted by $\Delta$), with the size $m - ws$, is first oversampled and then transferred to the LTM.

\subsection{Balanced representation of remaining set}
\label{sec:oversampling}
Before transferring the information from the STM to the LTM, we perform oversampling on the remaining set $\Delta$ to ensure a balanced representation of both classes. We denote the oversampled remaining set by $O_\Delta$. We use  \emph{BorderlineSMOTE}\footnote{We used the parameter values $k=5$, $m=5$ for the BorderlineSMOTE.}~\cite{han2005borderline}, which selects instances of the minority class that are misclassified by a kNN classifier and oversamples only those difficult instances, being more important for classification. The oversampled set is then transferred to the LTM.

\subsection{Transfer and cleaning:}
\label{sec:cleaning-transfer}
\begin{figure}[htbp]
\center
\includegraphics[width=0.32\textwidth]{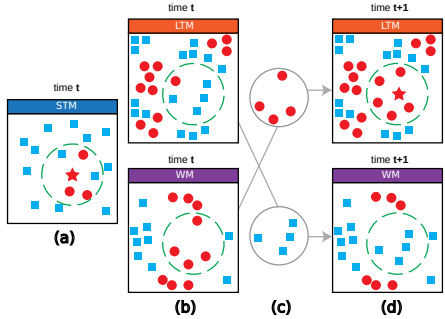}
\caption{Resolving the retroactive interference using a working memory incorporated into a multi-memory model. \textbf{(a)} state of the STM at time point $t$; the red star point denotes the most recent instance in the STM. \textbf{(b)} state of the LTM and WM at time point $t$; there are four \emph{inconsistent} instances in the LTM and four \emph{consistent} instances in the WM. \textbf{(c)} consistent instances (in red) and inconsistent instances (in blue) with respect to the most recent instance. \textbf{(d)} state of the LTM and WM at the time point $t+1$; the inconsistent instances in the LTM are exchanged with the consistent instances in the WM.} 
\label{fig:working_memory}
\end{figure}

DAM3 keeps the LTM consistent with the STM, similar to the SAM-kNN~\cite{losing2016knn}. The LTM is cleaned with respect to every incoming instance from the stream. However, DAM3 does not perform any cleaning on the remaining set (transferred from the STM to the LTM) due to its designed drift detector. SAM-kNN performs the cleaning as all the instances of the remaining set might not belong to the previous concept. In both cases (transfer of information from the STM to the LTM and cleaning of the LTM), the deletion of inconsistent instances might lead to the retroactive interference problem (c.f., \autoref{fig:retroactive_interference}).
To allow for the deletion of outdated information and also to prevent the deletion of older information due to the interference with newer information, we use the working memory. This memory resolves the retroactive interference by exchanging its consistent information with the inconsistent information of the LTM for every incoming instance from the stream. \autoref{fig:working_memory} illustrates how this problem is resolved.

We use the basic cleaning operation, proposed by the SAM-kNN~\cite{losing2016knn}, that deletes inconsistent instances of different classes in the neighborhood of an instance. The main idea is that most recent instances of the stream convey the correct class-label, and instances of different labels in its neighborhood should be considered as outdated and thus deleted.

\emph{Transfer of information from STM to LTM:}
The instances of $O_\Delta$ (oversampled remaining set) are transferred to the LTM and the $LTM$ is updated as follows:
\[
\begin{aligned}
LTM_{{t_d}+1}= LTM_{t_d} \cup O_{\Delta_{t_d}},
\end{aligned} 
\]
where $t_d$ is the time at which drift occurs.

\emph{Transfer of inconsistent information from LTM to WM:}
The $k$ nearest neighbors of $\mathbf{x}_{i}$ in $STM \backslash\left(\mathbf{x}_{i}, y_{i}\right)$, at time $t$, are determined and the ones with label $y_{i}$ are selected. The \emph{distance threshold} $\theta$ at time $t$ is then defined as:
\[
\begin{array}{c}
\theta_{t}=\max \left\{d\left(\mathbf{x}_{i}, \mathbf{x}\right)\right) |\; \mathbf{x} \in N_{k}\left(\mathbf{x}_{i}, STM_{t} \backslash\left(\mathbf{x}_{i}, y_{i}\right)\right), \\
\left.y(\mathbf{x})=y_{i}\right\}.
\end{array}
\]

On the basis of the found distance threshold of the $STM_{t}$, we define the \emph{inconsistent set} of the LTM at time $t$ ($IS_{LTM_{t}}$) with respect to the instance $\left(\mathbf{x}_{i}, y_{i}\right)$ in the $STM_{t}$:
\[
\begin{aligned}
IS_{LTM_{t}}=LTM_{t} \cap \left\{\left(\mathbf{x}_{j}, y\left(\mathbf{x}_{j}\right)\right) | \;\mathbf{x}_{j} \in N_{k}\left(\mathbf{x}_{i}, LTM_{t}\right)\right.,\\
\left.d\left(\mathbf{x}_{j}, \mathbf{x}_{i}\right) \leq \theta, \;y\left(\mathbf{x}_{j}\right) \neq y_{i}\right\}.
\end{aligned}
\]

Similarly, we define the \emph{consistent set} of the WM at time $t$ ($CS_{WM_{t}}$) with respect to the set $LTM_{t}$ and the instance $\left(\mathbf{x}_{i}, y_{i}\right)$ in the $STM$:

\[
\begin{aligned}
CS_{WM_{t}}=WM_{t} \cap \left\{\left(\mathbf{x}_{j}, y\left(\mathbf{x}_{j}\right)\right) | \;\mathbf{x}_{j} \in N_{k}\left(\mathbf{x}_{i}, WM_{t}\right)\right.,\\
\left.d\left(\mathbf{x}_{j}, \mathbf{x}_{i}\right) \leq \theta, \;y\left(\mathbf{x}_{j}\right) = y_{i}\right\}.
\end{aligned}
\]

Based on the $IS_{LTM_{t}}$ and $CS_{WM_{t}}$, inconsistent instances of the LTM ($IS_{LTM_{t}}$) are transferred to the WM and information of the $WM$ is updated as follows:
\[
\begin{aligned}
WM_{t+1}=\left(WM_{t} \backslash CS_{WM_{t}}\right) \cup IS_{LTM_{t}}.
\end{aligned}
\]
\emph{Transfer of consistent information from WM to LTM:} Consistent instances of the WM are transferred back to the LTM and information of the $LTM$ is updated as follows: 
\[
\begin{aligned}
LTM_{t+1}=\left(LTM_{t} \backslash IS_{LTM_{t}}\right) \cup CS_{WM_{t}}.
\end{aligned} 
\]
The WM preserves inconsistent information of the LTM, as it might be useful for later predictions, and makes its consistent information available to the LTM for current predictions.

\textbf{Compression of the LTM and WM.}
\label{sec:compression-ltm-wm}
The information of the LTM and WM is not discarded when their size exceeds the maximum threshold. Instead, as in SAM-kNN~\cite{losing2016knn}, we compress the data through class-wise \emph{K-Means++} clustering, such that the number of instances per class is reduced by half.

\subsection{Noise removal}
\label{sec:noise-removal}
The exchange of inconsistent instances of the LTM with the consistent instances of the WM might result in the generation of noisy instances. This means that some instances might become consistent with the LTM. We remove those instances which could be correctly classified based on the information of only the LTM, at each timepoint.

\section{Experiments}
\label{sec:experiments}
First, we compare DAM3 with state-of-the-art competitors on a variety of datasets (\autoref{sec:datasets}) using appropriate measures and evaluation setup (\autoref{sec:evaulation_setup}). The results are discussed in \autoref{sec:predictive_performance}.  Then, we focus on the behavior of our proposed model, namely the class-imbalance ratio of each memory (\autoref{sec:exp_behavior}), the interaction between the different memories, i.e., cleaning and transfer operations, (\autoref{sec:exp_cleaning_transfer}), and the size of memories with respect to the minority and majority classes (\autoref{sec:exp_memory_size}).

All the experiments were implemented and evaluated in Python using the scikit-multiflow framework \cite{montiel2018scikit}. The code and datasets are available on GitHub\footnote{https://amir-abolfazli.github.io/DAM3/}.

\label{sec:competitors}
For all the competitors, we used the default parameter values reported in the corresponding papers. The parameter values of the competitors, including \ours, are given in~\autoref{tab:parameterization}. We used the $\emph{pretrainSize} = 200$ for all classifiers.

\begin{table}[htbp!]
\caption{Parameter values for DAM3 and compared classifiers.}
\label{tab:parameterization}
  \centering
  \begin{adjustbox}{width=0.72\columnwidth}
\begin{tabular}{l l c l c}
    \toprule
    {Classifier} & {Parameter} & {Value} & {Parameter} & {Value}  \\ \midrule
    {DAM3} & {$n\_neighbors$}  & 5 & {$ltm\_size$}  & 0.5 \\
    {} & {$weighting$}  & distance & {$max\_window\_size$}  & 5000\\
    {} & {$min\_stm\_size$}  & 50 & {$drift\_detector\_ws$}  & min\_stm\_size \\
    {} & {$wm\_size$}  & 0.3& {$drift\_detector\_sig\_level$}  & 0.001 \\  

    \midrule
    {SAM-kNN~\cite{losing2016knn}} & {$n\_neighbors$}  & 5 & {$ltm\_size$}  & 0.4 \\
    {} & {$weighting$}  & distance & {$max\_window\_size$}  & 5000 \\
    {} & {$min\_stm\_size$}  & 50 & {} & {} \\
    \midrule
    {AdaC2~\cite{wang2016online}} & {$num\_estimators$}  & 10  & {$cost\_positive$}  & 1  \\
    {} & {$base\_estimator$}  & KNNAdwin & {$cost\_negative$}  & 0.1 \\
    {} & {$n\_neighbors$}  & 5 & {$drift\_detector$}  & Adwin \\

    \midrule
    {RUSBoost~\cite{wang2016online}} & {$num\_estimators$}  & 10 & {$algorithm$}  & 1 \\
    {} & {$base\_estimator$}  & KNNAdwin & {$drift\_detector$}  & Adwin \\
    {} & {$n\_neighbors$}  & 5 & {} & {} \\

    \midrule
    {LB~\cite{bifet2010leveraging}} & {$num\_estimators$}  & 10  & {$lambda$}  & 6 \\
    {} & {$base\_estimator$}  & kNN  & {$delta$} & 0.002 \\
    {} & {$n\_neighbors$}  & 5 & {$drift\_detector$}  & Adwin \\
    \midrule

    {OBA~\cite{oza2005online}} & {$num\_estimators$}  & 10  & {$n\_neighbors$}  & 5 \\
    {} & {$base\_estimator$}  & KNNAdwin  & {$drift\_detector$}  & Adwin \\
    \bottomrule
\end{tabular}
\end{adjustbox}
\end{table}

\subsection{Datasets}
\label{sec:datasets}
We experimented with a variety of synthetic and real datasets summarized in terms of their cardinality, dimensionality, class ratio, and drift type in~\autoref{tab:datasets}, as described below.

\textbf{Synthetic datasets:} Synthetic datasets have the advantage that any desired drift behavior can be explicitly simulated. We used the MOA framework \cite{bifet2010moa} to generate 4 synthetic streams with different types of concept drift. The SEA generator \cite{street2001streaming} was used to generate two streams: one stream with three sudden drifts and a constant 1:10 IR (\emph{SEA\_S}); and one stream with three gradual drifts where each concept has a different IR 1:4/1:5/1:2/1:10 (\emph{SEA\_G}). Similarly, the Hyperplane generator \cite{wang2003mining} was used to simulate two streams with incremental drifts: one with a constant IR 1:10 (\emph{HyperFast}); and one with a dynamic IR 1:1 $\rightarrow$ 1:100 (\emph{HyperSlow}). For the streams \emph{SEA\_S} and \emph{HyperFast} (constant $IR$=1:10), we included 10\% noise, and for the streams \emph{SEA\_G} and \emph{HyperSlow} (having different IRs and dynamic IR, respectively), we included 5\% noise.

\textbf{Real-world datasets:} 
Real-world datasets are used to show how well the stream classifiers perform in practice. A few  real-world  drift  benchmarks  are  available for binary classification, of  which we considered weather, electricity, and PIMA.

The \emph{weather} dataset \cite{elwell2011incremental} contains 18,159 instances and 8 features corresponding to the measurements such as temperature and wind speed. The goal is to predict whether it will be a rainy day (minority class) or not.

The \emph{electricity} dataset \cite{harries1999splice} contains 45,312 instances and 8 features such as date, demand, and price, and the goal is to predict whether the price will increase (minority class) or not, according to the moving average of last 24 hours.

The \emph{PIMA} Indian dataset \cite{smith1988using} contains 768 instances and 8 features, such as blood pressure, insulin, and age. The goal is to diagnostically predict whether a patient will have diabetes mellitus (minority class) or not, in 1-5 years.

\begin{table}[htbp!]
\caption{The characteristics of the datasets used in the
experiments.}
\label{tab:datasets}
  \centering
\begin{adjustbox}{width=0.75\columnwidth}
\begin{tabular}{l l c c c c c c}
    \toprule
    {Type} & {Datasets} & {\#Instances} & {\#Features} & {Class Ratio (+:-)} & {Noise} & {\#Drifts} & {Drift Type} \\ \midrule
    {Synthetic} & {SEA\_S} & 100K & 3 & 1:10 & 10\% & 3 & sudden \\
    {} & {SEA\_G}  & 100K & 3 & 1:4/1:5/1:2/1:10 & 5\% & 3 & gradual \\
    {} & {HyperFast}  & 50K & 5 & 1:10 & 10\% & 1 & incremental \\
    {} & {HyperSlow}  & 50K & 5 & 1:1 $\rightarrow$ 1:100 & 5\% & 1 & incremental \\
    \midrule
    {Real-world} & {Weather} & 18159 & 8 & 1:2.17 & {N/A} & {N/A} & {N/A} \\ 
    {} & {Electricity}  & 45312 & 8 & 1:1.35 & {N/A} & {N/A} & {N/A} \\
    {} & {PIMA} & 768 & 8 & 1:1.85 & {N/A} & {N/A} & {N/A} \\
    \bottomrule
\end{tabular}
\end{adjustbox}
\end{table}

\subsection{Evaluation setup}
\label{sec:evaulation_setup}
\subsubsection{Performance metrics}
\label{sec:measures}
An appropriate performance metric takes into account the performance on all classes, rather than the overall performance which is heavily affected by the majority class. The \emph{balanced accuracy} is an appropriate performance metric for imbalanced data. For binary classification, it is defined as the arithmetic mean of the sensitivity and specificity~\cite{fernandez2018learning}. Another performance metric, often used for imbalanced data, is the \emph{geometric mean (G-Mean)}. For binary classification, G-Mean is defined as the squared root of the product of the sensitivity and specificity. G-Mean punishes those models for which there is a big disparity between the sensitivity and specificity. It is different from the balanced accuracy which treats both classes equally~\cite{garcia2009index}.

\subsubsection{Evaluation method}
\label{sec:evaluation_method}
In data stream classification, the most commonly used evaluation method is the \emph{prequential evaluation}~\cite{gama2013evaluating}. The prequential evaluation is specifically designed for streaming settings, where instances arrive in sequential order. The idea is to first test the model on the instance, and then that instance is used to update the model. In this way, the model is always tested on the instances, not seen yet. The prequential evaluation is preferred over the traditional holdout evaluation as it makes the maximum use of the available data (i.e., no test set is needed) \cite{bifet2018machine}.

For the experiments, We evaluate the performance of the classifiers using the prequential evaluation and report on sensitivity, specificity, G-Mean, and balanced accuracy.

\subsection{Predictive performance}
\label{sec:predictive_performance}
\begin{figure}[!ht]
 \center
  \includegraphics[width=0.40\textwidth]{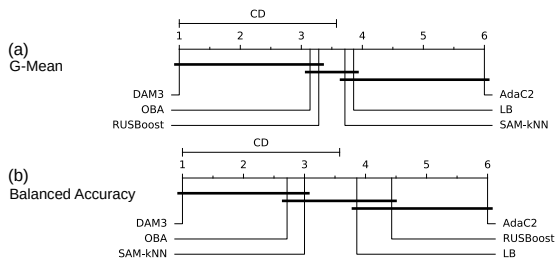}
  \caption{Critical difference diagram for the post-hoc Bonferroni-Dunn test.}
  \label{fig:cd_diagram}
\end{figure}

In~\autoref{tab:predictiveperformanceoverview}, the predictive performance of DAM3 and compared classifiers on the different datasets is shown. 
The proposed method DAM3 outperforms all the compared methods in terms of G-Mean and balanced accuracy. This indicates that DAM3 finds a trade-off between the sensitivity and specificity better than other methods. To further investigate the differences in the average G-Mean and balanced accuracy (i.e., average ranks) of the compared methods on the considered datasets, we used the post-hoc Bonferroni-Dunn test~\cite{demvsar2006statistical} to compute the critical difference (CD). The results are shown in \autoref{fig:cd_diagram} and as we can see, the performance of DAM3 is significantly better than AdaC2, LB, and SAM-kNN in terms of G-Mean, and significantly better than AdaC2, RusBoost, and LB, in terms of balanced accuracy.

\begin{table}[htbp!]
  \caption{Predictive performance of the classifiers. Classifiers with the best and second-best performance are marked in bold and underlined, respectively.}
  \label{tab:predictiveperformanceoverview}
  \centering
  \resizebox{0.72\columnwidth}{!}{%
\begin{tabular}{l l c c c c c c}
    \toprule
    {Dataset} & {Classifier} & {Sensitivity} & {Specificity} & {G-Mean} & {Balanced Accuracy} \\ \midrule
    {SEA\_S} & {DAM3}  & \underline{0.4614} & 0.9190 & \textbf{0.6512} & \textbf{0.6902} \\
    {} & {SAM-kNN}  & 0.3778 & \textbf{0.9884} & 0.6111 & \underline{0.6831} \\
    {} & {AdaC2} & 0.1449 & \underline{0.9883} & 0.3783 & 0.5666 \\
    {} & {RUSBoost}  & \textbf{0.4777} & 0.7961 & \underline{0.6167} & 0.6369 \\
    {} & {LB} & 0.3923 & 0.8826 & 0.5884 & 0.6374 \\
    {} & {OBA}  & 0.3845 & 0.9795 & 0.6137 & 0.6820 \\
    \midrule
    {SEA\_G} & {DAM3}  & \textbf{0.6504} & 0.9393 & \textbf{0.7816} & \textbf{0.7946} \\
    {} & {SAM-kNN} & 0.5467 &  \textbf{0.9919} & 0.7364 & 0.7693 \\
    {} & {AdaC2} & 0.2843 & \underline{0.9913} & 0.5308 & 0.6378 \\
    {} & {RUSBoost}  & 0.5417 & 0.8523 & 0.6794 & 0.6970 \\
    {} & {LB} & 0.5307 & 0.9345 & 0.7043 & 0.7326 \\
   {} & {OBA} & \underline{0.5569} & 0.9901 & \underline{0.7426} & \underline{0.7735} \\
    \midrule
    {HyperFast} & {DAM3} & \underline{0.4026} & 0.9021 & \textbf{0.6026} & \textbf{0.6523} \\
    {} & {SAM-kNN}  & 0.2205 & \textbf{0.9930} & 0.4680 & 0.6067 \\
    {} & {AdaC2} & 0.0912 & \underline{0.9900} & 0.3005 & 0.5406 \\
    {} & {RUSBoost} & \textbf{0.4144} & 0.7883 & \underline{0.5716} & 0.6014 \\
    {} & {LB} & 0.2976 & 0.8846 & 0.5131 & 0.5911 \\
   {} & {OBA}  & 0.2596 & 0.9809 & 0.5046 & \underline{0.6202} \\
    \midrule
    {HyperSlow} & {DAM3}  & \textbf{0.3990} & 0.9230 & \textbf{0.6069} & \textbf{0.6610} \\
    {} & {SAM-kNN}  & 0.2852 &  \textbf{0.9960} & 0.5330 & \underline{0.6406} \\
    {} & {AdaC2} & 0.1180 & 0.9942 & 0.3426 & 0.5561 \\
    {} & {RUSBoost} &  \underline{0.3768} & 0.8728 & \underline{0.5734} & 0.6248 \\
    {} & {LB} & 0.2795 & 0.9428 & 0.5133 & 0.6112 \\
   {} & {OBA}  & 0.2668 & \underline{0.9949} & 0.5152 & 0.6308 \\
    \midrule
    {Weather} & {DAM3}  & \underline{0.7479} & 0.7663 & \textbf{0.7570} & \textbf{0.7571} \\
    {} & {SAM-kNN} & 0.5080 & \textbf{0.9068} & 0.6788 & 0.7074 \\
    {} & {AdaC2} & \textbf{0.8982} & 0.4211 & 0.6150 & 0.6596 \\
    {} & {RUSBoost}  & 0.5549 & 0.7869 & 0.6608 & 0.6709 \\
    {} & {LB} & 0.5501 & 0.8052 & 0.6655 & 0.6777 \\
    {} & {OBA}  & 0.5627 & \underline{0.8785} & \underline{0.7031} & \underline{0.7206} \\
    \midrule
    {Electricity} & {DAM3}  & \textbf{0.8335} & \underline{0.8780} & \textbf{0.8555} & \textbf{0.8558} \\
    {} & {SAM-kNN} & 0.7879 &  0.8566 & 0.8215 & 0.8222 \\
    {} & {AdaC2} & 0.3824 & \textbf{0.9681} & 0.6085 & 0.6753 \\
    {} & {RUSBoost}  & 0.8017 & 0.8214 & 0.8115 & 0.8115 \\
    {} & {LB} & \underline{0.8233} & 0.8709 & \underline{0.8468} & \underline{0.8471} \\
    {} & {OBA}  & 0.7274 & 0.8285 & 0.7763 & 0.7780 \\
    \midrule
    {PIMA} &  {DAM3}  & \underline{0.7143} & 0.7079 & \textbf{0.7111} & \textbf{0.7111}\\
    {} & {SAM-kNN} & 0.3308 &  \underline{0.8727} & 0.5373 & 0.6017 \\
    {} & {AdaC2} & 0.2247 & \textbf{0.9699} & 0.4669 & 0.5346 \\
    {} & {RUSBoost}  & 0.4962 & 0.7228 & 0.5989 & 0.6095 \\
    {} & {LB} & \textbf{0.7491} & 0.5113 & 0.6189 & 0.6302 \\
    {} & {OBA}  & 0.5489 & 0.8090 & \underline{0.6664} & \underline{0.6789} \\
    \bottomrule
\end{tabular}
}
\end{table}

\textbf{Ablation study.} In DAM3, all the components (drift detector, oversampling, working memory, and weighting of classifiers based on balanced accuracy) work together to mitigate the class imbalance. Therefore, showing the impact of each component alone does not make sense. Apart from these components, one of the main differences between DAM3 and SAM-kNN is the use of full Bayes as the STM classifier (instead of the weighted kNN). 

In \autoref{tab:predictiveperformance_FB}, we compare the performance of DAM3 in terms of G-Mean and balanced accuracy with the performance of SAM-kNN with both kNN and full Bayes as the STM classifier, and show the difference in the performance. The results indicate that the use of the full Bayes as the STM classifier in SAM-kNN slightly improves the performance on all datasets (except SEA\_S and Weather). The results also show that for all the datasets (except Electricity), the superior performance of DAM3 is mainly due to the considered components and not the use of the full Bayes classifier.
\begin{table}[htbp!]
  \caption{Impact of the full Bayes classifier used as STM classifier for SAM-kNN, compared with the original SAM-kNN and DAM3.}
  \label{tab:predictiveperformance_FB}
  \centering
  \resizebox{0.84\columnwidth}{!}{%
\begin{tabular}{l l c c c c c c}
    \toprule
    {Dataset} & {Classifier} & {G-Mean} & {G-Mean Diff} & {Balanced Accuracy} & {Balanced Accuracy Diff} \\ \midrule
    {SEA\_S} & {$\mathrm{DAM3}$} & 0.6512 & \textcolor{teal}{$\uparrow$ 0.0457} & 0.6902 & \textcolor{teal}{$\uparrow$ 0.0103} \\
    \cmidrule(l){2-6}
    {} & {$\mathrm{SAMkNN-STM\_{FB}}$}  & 0.6055 & \textcolor{red}{$\downarrow$ 0.0056}  &  0.6799  &  \textcolor{red}{$\downarrow$ 0.0032}\\
    {} & {$\mathrm{SAMkNN-STM\_{kNN}}$}  & 0.6111 & &   0.6831  &  \\
    \midrule
    {SEA\_G} & {$\mathrm{DAM3}$} & 0.7816 & \textcolor{teal}{$\uparrow$ 0.0364} &  0.7946 & \textcolor{teal}{$\uparrow$ 0.0185} \\
    \cmidrule(l){2-6}
    {} & {$\mathrm{SAMkNN-STM\_{FB}}$}  & 0.7452 & \textcolor{teal}{$\uparrow$ 0.0088}   &  0.7761 &   \textcolor{teal}{$\uparrow$ 0.0068} \\
    {} & {$\mathrm{SAMkNN-STM\_{kNN}}$}  & 0.7364 & &   0.7693 &   \\
    \midrule
    {HyperFast} & {$\mathrm{DAM3}$}  & 0.6026 & \textcolor{teal}{$\uparrow$ 0.0957} & 0.6523 & \textcolor{teal}{$\uparrow$ 0.0290} \\
    \cmidrule(l){2-6}
    {} & {$\mathrm{SAMkNN-STM\_{FB}}$}  & 0.5069 & \textcolor{teal}{$\uparrow$ 0.0389}  &  0.6233  &  \textcolor{teal}{$\uparrow$ 0.0166} \\
    {} & {$\mathrm{SAMkNN-STM\_{kNN}}$} & 0.4680 &  &  0.6067  &  \\
    \midrule
    {HyperSlow} & {$\mathrm{DAM3}$} & 0.6069 & \textcolor{teal}{$\uparrow$ 0.0494} &  0.6610  & \textcolor{teal}{$\uparrow$ 0.0076} \\
    \cmidrule(l){2-6}
    {} & {$\mathrm{SAMkNN-STM\_{FB}}$}  & 0.5575 & \textcolor{teal}{$\uparrow$ 0.0245}  &  0.6534  &  \textcolor{teal}{$\uparrow$ 0.0128} \\
    {} & {$\mathrm{SAMkNN-STM\_{kNN}}$} & 0.5330 &  & 0.6406  &  \\
    \midrule
    {Weather} & {$\mathrm{DAM3}$}  & 0.7570 & \textcolor{teal}{$\uparrow$ 0.0828} &  0.7571 & \textcolor{teal}{$\uparrow$ 0.0528} \\
    \cmidrule(l){2-6}
    {} & {$\mathrm{SAMkNN-STM\_{FB}}$}  & 0.6742 & \textcolor{red}{$\downarrow$ 0.0046}  &  0.7043  &  \textcolor{red}{$\downarrow$ 0.0031} \\
    {} & {$\mathrm{SAMkNN-STM\_{kNN}}$}  & 0.6788 &  &  0.7074  &  \\
    \midrule
    {Electricity} & {$\mathrm{DAM3}$}  & 0.8555 & \textcolor{teal}{$\uparrow$ 0.0137} & 0.8558 & \textcolor{teal}{$\uparrow$ 0.0133} \\
    \cmidrule(l){2-6}
    {} & {$\mathrm{SAMkNN-STM\_{FB}}$}  & 0.8418 & \textcolor{teal}{$\uparrow$ 0.0203}  &  0.8425  &  \textcolor{teal}{$\uparrow$ 0.0203} \\
    {} & {$\mathrm{SAMkNN-STM\_{kNN}}$} & 0.8215 &  &  0.8222  &  \\
    \midrule
    {PIMA} & {$\mathrm{DAM3}$} & 0.7111 & \textcolor{teal}{$\uparrow$ 0.0986} & 0.7111  & \textcolor{teal}{$\uparrow$ 0.0698} \\
    \cmidrule(l){2-6}
    {} & {$\mathrm{SAMkNN-STM\_{FB}}$} & 0.6125 & \textcolor{teal}{$\uparrow$ 0.0752} &   0.6413 &   \textcolor{teal}{$\uparrow$ 0.0396} \\
    {} & {$\mathrm{SAMkNN-STM\_{kNN}}$} & 0.5373 &  &  0.6017 &   \\
    \bottomrule
\end{tabular}
}
\end{table}

\subsection{Model behavior}
The goal of this section is to shed light on the internal mechanisms of DAM3 and their contribution towards tackling both class imbalance and concept drifts.

\subsubsection{Imbalance perception by the model}
\label{sec:exp_behavior}
In this section, we examine the IR of the memories of our model, DAM3, compared with the cumulative IR of the stream. We also examine the IR of the memories of the SAM-kNN~\cite{losing2016knn} which is the most similar model to our model in terms of architecture.

\begin{figure}[!htbp]
 \center
  \includegraphics[width=0.38\textwidth]{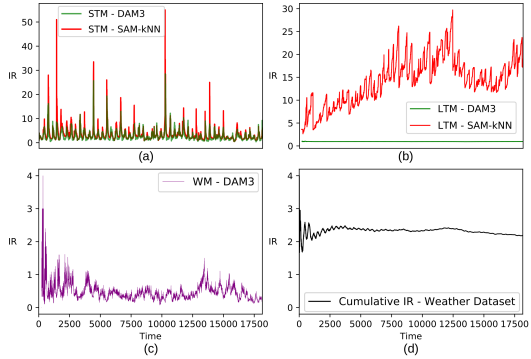}
  \caption{The IR of the memories of DAM3 compared with that of SAM-kNN on the Weather dataset.}
  \label{fig:weatherIR}
\end{figure}

\autoref{fig:weatherIR} (a) shows the IR of all memories for our model DAM3, compared with SAM-kNN, on the Weather dataset. The drift detector helps the model to not make the STM full with majority instances. As a result, the IR of the STM in DAM3 is lower than that of SAM-kNN. In (b), the green and red lines correspond to the IRs of the LTM for DAM3 and SAM-kNN, respectively. The lines reveal that what is reflected by SAM-kNN is significantly higher than the actual IR shown in (d). Moreover, for SAM-KNN, the IR of the LTM gradually increases over time, implying that the majority class becomes increasingly dominant in the LTM. Unlike SAM-kNN, DAM3 reflects a balanced representation of classes ($\mathrm{IR}\approx 1$). In (c), the purple line shows the IR of the WM that corresponds to the instances which are inconsistent with the LTM. 

\subsubsection{Removal and transfer of instances in LTM and WM}
\label{sec:exp_cleaning_transfer}
In this section, we demonstrate the ``removal'' and  ``transfer'' of information within the long-term and working memories.

\begin{figure}[!ht]
 \center
  \includegraphics[width=0.34\textwidth]{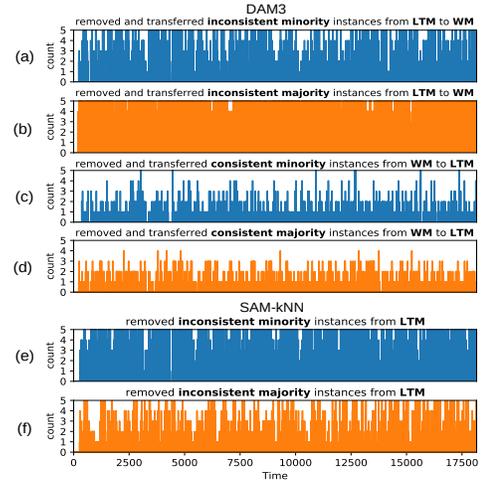}
  \caption{The number of removed and transferred instances from the LTM to WM and vice versa for DAM3, compared with the number of removed inconsistent instances from LTM for SAM-kNN, on the Weather dataset.}
  \label{fig:weather_memoryInconsistency}
\end{figure}

In \autoref{fig:weather_memoryInconsistency}, (a) and (b) show the number of \emph{inconsistent} minority and majority instances, removed and transferred from the LTM to the WM in DAM3. (c) and (d) show the number of \emph{consistent} minority and majority instances, removed and transferred from the WM to the LTM in DAM3. (e) and (f) show the number of \emph{inconsistent} minority and majority instances which are removed from the LTM in SAM-kNN. Since we used the kNN with k=5, at most, there could be 5 inconsistent instances to be removed and transferred at each time point. Comparing the subfigures (a) and (b) with (e) and (f) shows that DAM3 removes fewer minority instances compared with SAM-kNN. This statement is supported by the subfigure (b) in \autoref{fig:weatherIR}, showing the IR of the LTM (in red), where the IR increases gradually. This implies that SAM-kNN removes more minority instances over time. The subfigures (c) and (d) show the number of removed and transferred \emph{consistent} instances of the WM for both minority and majority classes. Both (c) and (d) show a similar behavior, revealing that there are some consistent instances which could be transferred back almost all the time. This means that the DAM3 resolves the problem of retroactive interference, impeding the model’s ability to retrieve the old minority instances. 

\subsubsection{Size of memories}
\label{sec:exp_memory_size}
\begin{figure}[!htpb]
 \center
  \includegraphics[width=0.47\textwidth]{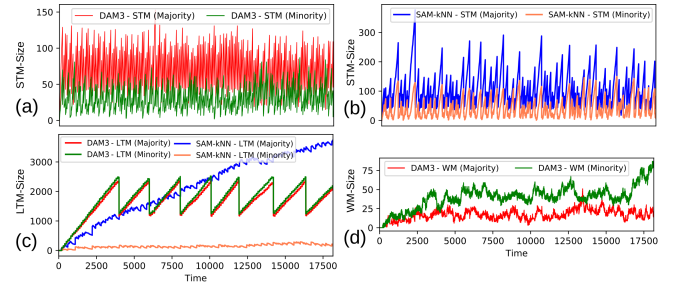}
  \caption{The size of all memories with respect to the minority and majority classes for both models DAM3 and SAM-kNN, on the Weather dataset.}
  \label{fig:memory_size}
\end{figure}

In \autoref{fig:memory_size}, (a) and (b) correspond to the size of STM with respect to the minority and majority for the models DAM3 and SAM-kNN, respectively. Both subfigures show a similar behavior, however, the size of STM in DAM3 is, on average, 32\% smaller than that of the SAM-kNN, due to the use of the drift detector. In (c), the number of majority instances in the LTM, for the SAM-kNN (blue line), gradually increases while the number of minority instances (red line) remains almost the same. The behavior is completely different for DAM3, where the number of minority instances is almost equal to the number of majority instances. The sudden drops in the size of the LTM correspond to the times at which compression occurs. In (d), green and red lines correspond to the number of minority and majority instances, respectively, in the WM.

\section{Conclusion}
\label{sec:conclusion}
In this paper, we proposed the Drift-Aware Multi-Memory Model (DAM3), designed to mitigate the class imbalance in dual-memory models, dedicating short-term and long-term memories to the current and former concepts, respectively. DAM3 mitigates the class imbalance by incorporating an imbalance-sensitive drift detector, preserving a balanced representation of classes in the long-term memory, resolving the retroactive interference using a working memory preventing the removal of old information, and weighting the classifiers induced on different memories based on their balanced accuracy. Our experimental results showed that the proposed method outperforms the state-of-the-art methods in terms of G-Mean and balanced accuracy. For future work, we intend to design a multi-memory model that deals with recurring 
drifts.

\section*{Acknowledgment}
The work of the first author was supported by the German Research
Foundation (DFG) within the project OSCAR (Opinion Stream Classification with Ensembles and Active leaRners) and HEPHAESTUS (Machine learning methods for adaptive process planning of 5-axis milling), for both of which the second author is a principal investigator. 
\bibliographystyle{IEEEtran}
\bibliography{main}
\end{document}